\documentclass[letterpaper, 10 pt, conference]{ieeeconf}  

\IEEEoverridecommandlockouts                              

\usepackage{graphicx} 
\graphicspath{ {./images/} }

\usepackage[rightcaption]{sidecap}

\usepackage{wrapfig}
\usepackage{gensymb}

\usepackage{siunitx}
\usepackage{xcolor}
\usepackage{float}

\usepackage{multirow}
\usepackage{tabu}
\usepackage{subfigure} 
\usepackage{amssymb}
\usepackage{balance}
\usepackage{algorithm}
\usepackage{algorithmic}
\usepackage{amsmath}

\DeclareMathOperator*{\argmax}{argmax}

\title{\LARGE \bf
Dynamic Humanoid Locomotion over Uneven Terrain \\
With Streamlined Perception-Control Pipeline
}

\author{
Moonyoung Lee$^{1}$, Youngsun Kwon$^{2}$, Sebin Lee$^{2}$, JongHun Choe$^{1}$, Junyong Park$^{3}$, \\
Hyobin Jeong$^{4}$, Yujin Heo$^{1}$,  Min-su Kim$^{1}$, Jo Sungho$^{3}$, Sung-Eui Yoon$^{2}$, Jun-Ho Oh$^{1}$
\thanks{$^{1}$Is with the Humanoid Robot Research Center, Department of
Mechanical Engineering, Korea Advanced Institute of Science and Technology, Daejeon, Korea 
        {\tt\small jhoh@kaist.ac.kr}}%
\thanks{$^{2}$Is with the Scalable Graphics, Vision \& Robotics Lab, School of Computing, Korea Advanced Institute of Science and Technology, Korea}
\thanks{$^{3}$Is with the Neuro-Machine Augmented Intelligence Laboratory, School of Computing, Korea Advanced Institute of Science and Technology, Korea}
\thanks{$^{4}$Is with the Korea Atomic Energy Research Institute (KAERI), Korea}
}

\begin{document}

\maketitle
\thispagestyle{empty}
\pagestyle{empty}

\begin{abstract}

Although bipedal locomotion provides the ability to traverse unstructured environments, it requires careful planning and control to safely walk across without falling. This poses an integrated challenge for the robot to perceive, plan, and control its movements, especially with dynamic motions where the robot may have to adapt its swing-leg trajectory on-the-fly in order to safely place its foot on the uneven terrain.

In this paper we present an efficient geometric footstep planner and the corresponding walking controller that enables a humanoid robot to dynamically walk across uneven terrain at speeds up to 0.3 m/s. As dynamic locomotion, we refer first to the continuous walking motion without stopping, and second to the on-the-fly replanning of the landing footstep position in middle of the swing phase during the robot gait cycle.
This is mainly achieved through the streamlined integration between an efficient sampling-based planner and robust walking controller. The footstep planner is able to generate feasible footsteps within 5 milliseconds, and the controller is able to generate a new corresponding swing leg trajectory as well as the whole-body motion to dynamically balance the robot to the newly updated footsteps.  
The proposed perception-control pipeline is evaluated and demonstrated with real experiments using a full-scale humanoid to traverse across uneven terrains featured by static stepping stones, 
dynamically movable 
stepping stone, or narrow path.

\end{abstract}


\section{INTRODUCTION}

A primary benefit of a bipedal design over tradition wheeled platform is that humanoid robots have potential for great mobility similar to that of humans. This mobility is important because it enables robots to more easily adapt to various real-world environments, 
e.g., stairs, stepping stones, and narrow paths,
that are already human-centric designed to accommodate legged locomotion.

Such extended maneuverability opens the possibility for bipedal robots to partake in wider range of applications such as in disaster response, in which traditional wheeled platforms are limited in terms of mobility. 
However, these real-world environments are typically unstructured and previously unseen by the robot. Furthermore, the environment the robot must traverse could be composed of uneven surfaces and unstable terrain. Such challenge entails that 
the robot must have highly dynamic performances in perception and control. 
More specifically, the robot must be able to perceive, plan, and react at a fraction of a second on a given terrain without falling down. 

In this work, we approach the problem 
based on an efficient sampling-based geometric planner for generating feasible footsteps, and a robust walking controller that can handle dynamic footstep updates in real-time.
There are many existing planning algorithms for navigating unknown environments, which can be divided into either global-map based algorithms or memoryless algorithms. To achieve faster perception and therefore, more dynamic performance, our method forgoes maintaining a global-map in favor of using only the most recent depth data in the local frame. We opted for this design decision as there are rarely cases where the robot has to walk in a direction that is not perceived in its current field of view, such as walking backwards, which would then require storing a global-map.   

\begin{figure}
    \centering
    \includegraphics[scale=0.7]{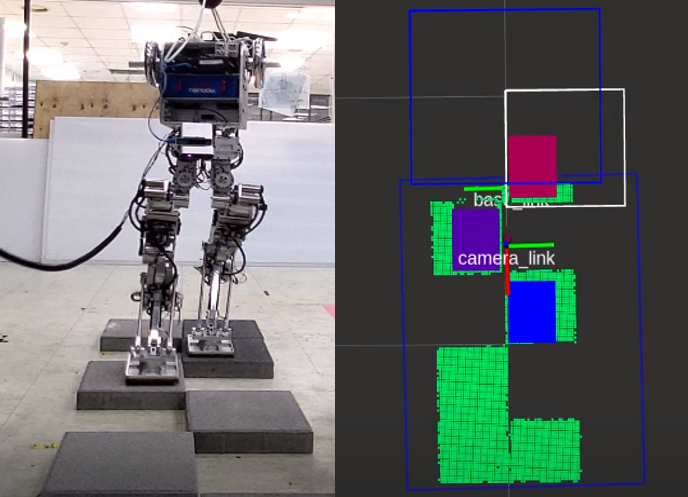}
    \caption{Terrain reconstruction and footstep placements dynamically generated to traverse uneven terrain. Color gradient represents the footstep sequence, starting with the immediate next step from robot base.}
    \label{fig:title_photo}
\end{figure}

Overall, our main contributions are
\begin{itemize}
    \item Streamlined integrated mapping, planning, and control pipeline for continuous legged locomotion over uneven terrain at speeds of 0.3 m/s using only on-board sensors, which is faster than comparable published works to the best of the author’s knowledge.
    \item Efficient sampling-based footstep planner that uses a proposed geometry-aware safety scoring to traverse uneven terrains. 
    \item Dynamic on-the-fly replanning of the landing footstep position in middle of the swing phase during the robot  gait cycle in order to safely traverse changing terrain.
    \item Extensive experimental results of the proposed perception-control pipeline using a full-sized humanoid in various environments featured by stepping stones, dynamically 
   movable stepping stones, or narrow path.  
\end{itemize}

The remaining of the paper is structured as follows. 
Section II: related works and comparison to our approach.
Section III: robot platform's software framework. Section IV: perception for mapping.
Section V: footstep planner and safety scoring. Section VI: foot trajectory generation and posture controller. 
We conclude with experimental results and discussion in Section VI.  
\section{RELATED WORK}

Impressive implementation of online footstep planning based on on-board vision sensors has been demonstrated by a number of research groups. For example, one notable forerunner in this field implemented a footstep planner for the Honda ASIMO robot based on 2D A* search to navigate dynamic environments~\cite{asimo_cmu}. This research could efficiently plan up to 30 steps in a second but utilized off-board motion capture system and distinct colored panels to simplify the perception pipeline. The work of ~\cite{2d_edge_detect} successfully implemented real-time perception and planning all on-board for stair-like terrains using stereo images by simplifying planar regions to  straight edges with a line detector. 

Our footstep planner resembles the geometric approach studied in ~\cite{2d_geometric_plan}. Karkowski et al. utilized RGB-D camera to dynamically adapt to local changes in the environment, and could generate feasible footstep path in 18 milliseconds. This planner, however, differs from our work by generating a 2D path to the global goal first, and then individual footsteps along the path, mainly focusing on avoiding obstacles on the flat ground. In contrast to navigating towards the global goal, the renown research of Fallon et al. proposed continuous walking autonomy using short-horizon footstep planning within safe regions to enable robot traversing forward over uneven terrain~\cite{mit_planning}. Their approach, rather than discrete graph-based search, used continuous search space algorithm based on MIQP leading to global optimal but long planning time of over 400 msec, which significantly reduced the overall execution-speed of the perception-action pipeline. 

The research of ~\cite{ihmc_planning} is most similar to ours in that it utilizes discrete search technique with heuristically scored nodes over the generated 3D map of the terrain. 
Though the aforementioned techniques showed various planning for humanoid locomotion in order to avoid obstacles or traverse complex terrains, there is no modification during the leg swing phase once a feasible footstep is generated. A different strategy is required to handle the case where the terrain of the landing foot suddenly changes due to instability or disturbances in the terrain. Our approach differs in that our pipeline allows for on-the-fly replanning and dynamically updating the landing footstep during the robot’s swing-leg phase. A very recent work~\cite{inaba_vision_recovery} similarly showed highly dynamic performances to quickly generate footsteps upon push-recoveries on uneven terrain. This work simplified the visual processing by utilizing color features to determine steppable regions. 

Our footstep planner generates feasible footsteps determined as best effort within a fixed sample time. Best effort here refers to scoring possible footstep array candidates, prioritizing candidates with a higher number of  footsteps sampled, and safety score based on distance from the edges of the terrain.
The footstep positions, starting with the next target foothold, is then updated in the Step Data Base of the controller, which generates a new swing leg trajectory as well as a whole-body motion to balance the robot as it reaches the updated footstep position. 

\newcommand{\figref}[1]{Figure.~\ref{#1}}
\newcommand{\tabref}[1]{Table~\ref{#1}}
\newcommand{\eqnref}[1]{Eqn.~(\ref{#1})}
\newcommand{\secref}[1]{Sec.\ref{#1}}
\newcommand{\ie}{{\it i.e.}}
\newcommand{\eg}{{\it e.g.}}
\renewcommand{\vec}[1]{\mathbf{#1}}



\section{System Overview}

In this work, we use the Gazelle legged platform~\cite{gazelle_design}, 
which is
a lightweight 13-DOF bipedal robot that is capable of walking at relatively high speeds with 30 cm stride with 0.5 sec step time (0.6 m/s).
Our previous work with this platform focused on dynamic walk without perception capability~\cite{hyobin_tro} or on push recoveries on flat grounds~\cite{gazelle_vision}. This work extends robust walking controller with the proposed footstep planner to traverse uneven terrain.


\begin{figure}[t]
    \centering
    \includegraphics[scale=0.7]{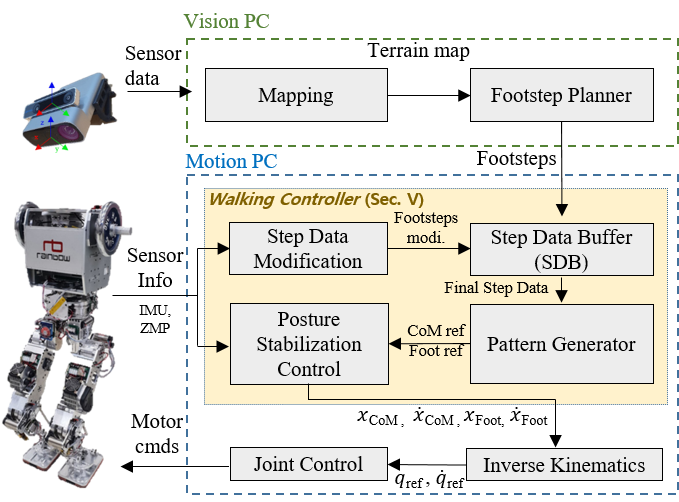}
    \caption{Overall software diagram illustrating the flow of modules and signals through system}
    \label{fig:sw_diagram}
\end{figure}

The software framework is intuitively divided among two PCs (6th gen i7, 4 cores) for vision and motion control as shown in Figure~\ref{fig:sw_diagram}. Vision PC utilizes ROS to handle camera data, such as the Intel T265 for high-speed VIO at 200Hz, and the Microsoft Kinect Azure for time-of-flight depth camera at 30Hz. From these camera data, a local 3D octomap of the terrain is generated with 1cm resolution of rates up to 15 Hz. During the single support phase of the gait cycle, which has least effect on the cameras from walking vibrations, the next array of feasible footstep positions is requested and generated within 5 msec. 

The Motion PC runs on the custom PODO software framework to leverage real-time performance, with whole-body motion controller iterating at 500 Hz to accommodate the generated footstep data. This integration of ROS-PODO framework~\cite{Mhubo}, which was previously designed for a wheeled robot in our previous work, is modified for interfacing with a legged robot.


\section{Terrain Mapping}

\begin{figure}[b]
    \centering
    \includegraphics[scale=0.65]{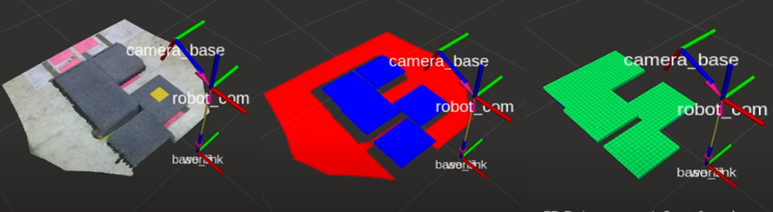}
    \caption{Overall procedure of obtaining the terrain map. Raw point cloud of the terrain (left),  major planes segmented using RANSAC (middle), 3D octomap constructed of the traversable region (right).}
    \label{fig:mapping}
\end{figure}

The terrain map is regenerated at every frame to allow fast updates of the current scene. Instead of using Simultaneous Localization and Mapping (SLAM) to create a global map expanding over the whole environment, we focus on creating a local map  directly in front of the robot. This strategy handles 
the drifting error, which is a common problem of many mapping algorithms. Furthermore, generating a real-time local map at every frame allows the map to be sensitive to dynamic changes in the environment.

Fankhauser et al.~\cite{fankhauser} propose a probabilistic terrain mapping method for quadruped robots where errors such as drifts and noise in the map are managed by formulating confidence levels in the map. However, their method is not appropriate for bipedal robots since it does not detect flat planes for footstep planning, and they show a much slower update rate for dynamic environments. Our map generation method is similar to the work of~\cite{ihmc_planning} in that planar surfaces are generated from the point cloud. 
While \cite{ihmc_planning} generates a complete map of the whole scene prior to the traversal,  we focus on generating a real-time local map at every frame. This 
enables re-planning of footsteps during the traversal. 

The overall procedure of our terrain mapping process is shown in Fig.~\ref{fig:mapping}. The raw point cloud of the terrain, obtained through the camera, is first transformed to the robots base frame to fit the orientation of the world. The point cloud is then down-sampled and cropped to remove points outside the region of interest. We choose a 1 meter wide and 2 meter long horizontal box as our default region of interest.

These parameters are adjusted to fit the resolution and computation time trade-off depending on the traversing environment. RANSAC is used to segment the major planes in the point cloud. Planes outside an angular range of the plane of the floor are removed to ignore non-terrain planes such as walls. Finally, the planes classified as traversable are chosen and converted into an octomap for the planner.

\begin{figure*}[t]
    \centering
    \subfigure [Footstep sampling]	                            { \includegraphics[height=1.5in]{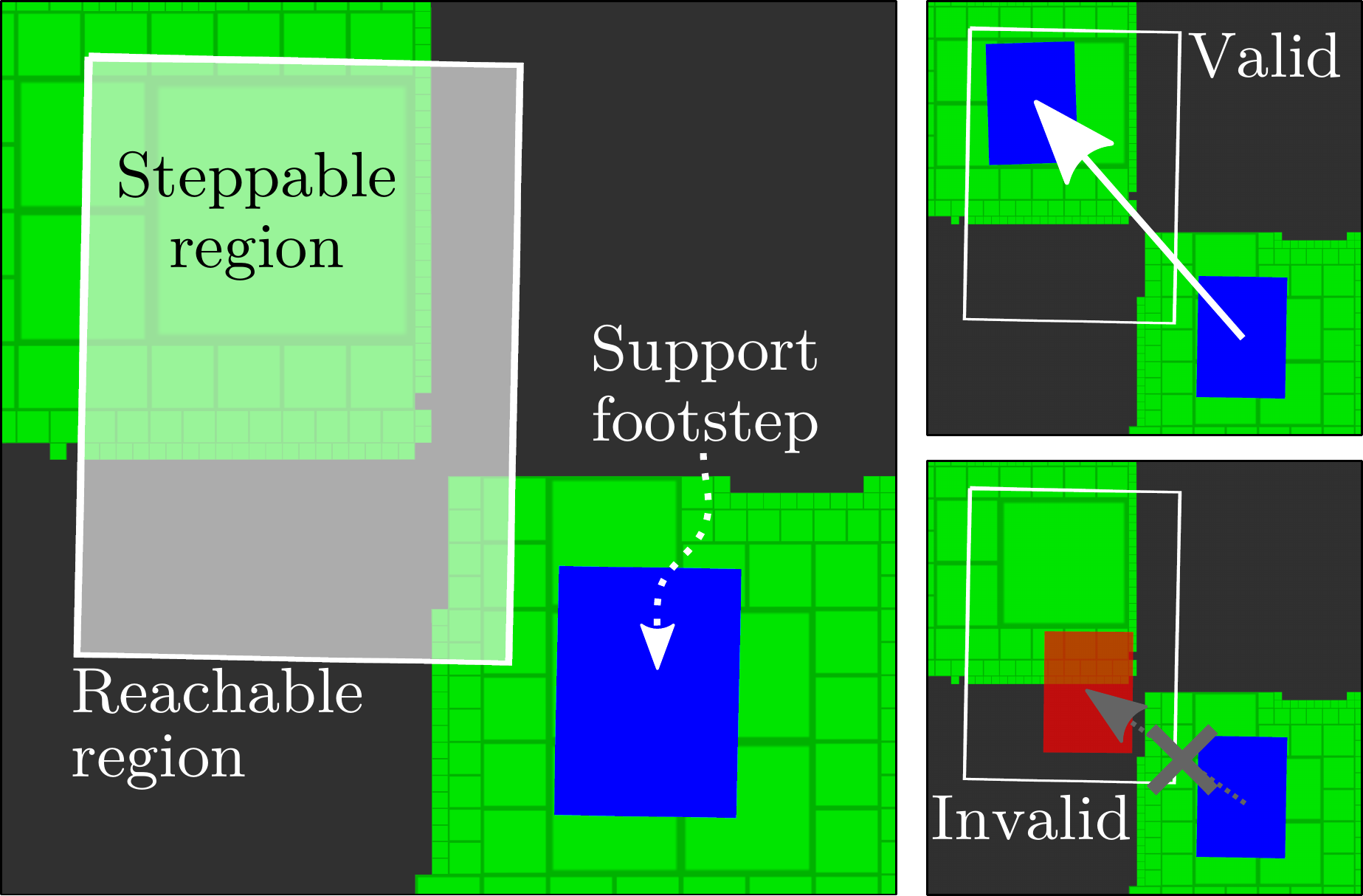} }
    \hspace{0.1cm}
    \subfigure [Footstep paths with different safety scores]	{ \includegraphics[height=1.5in]{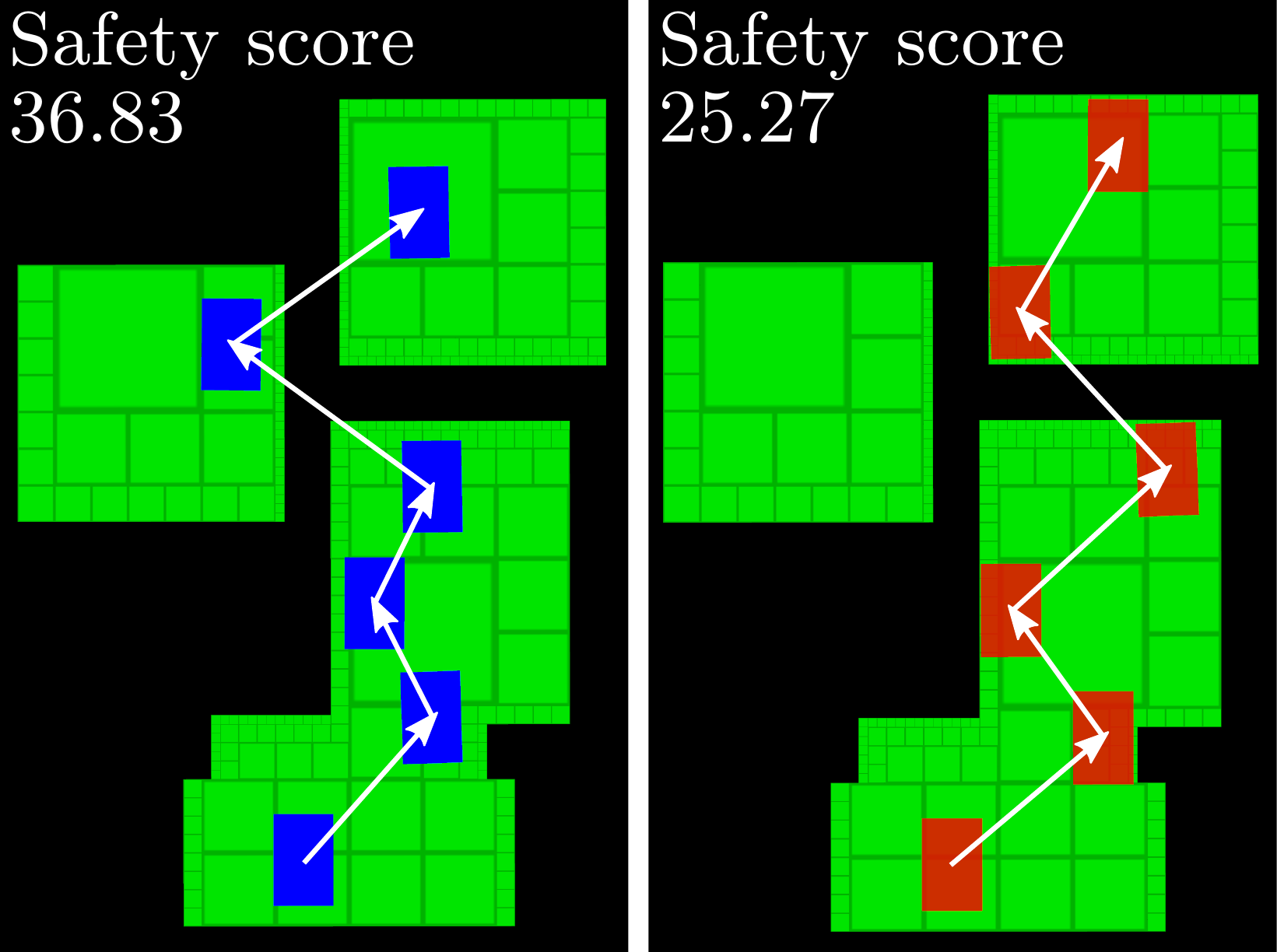} }
    \hspace{0.1cm}
    \subfigure [Test points for computing a safety score]	    { \includegraphics[height=1.5in]{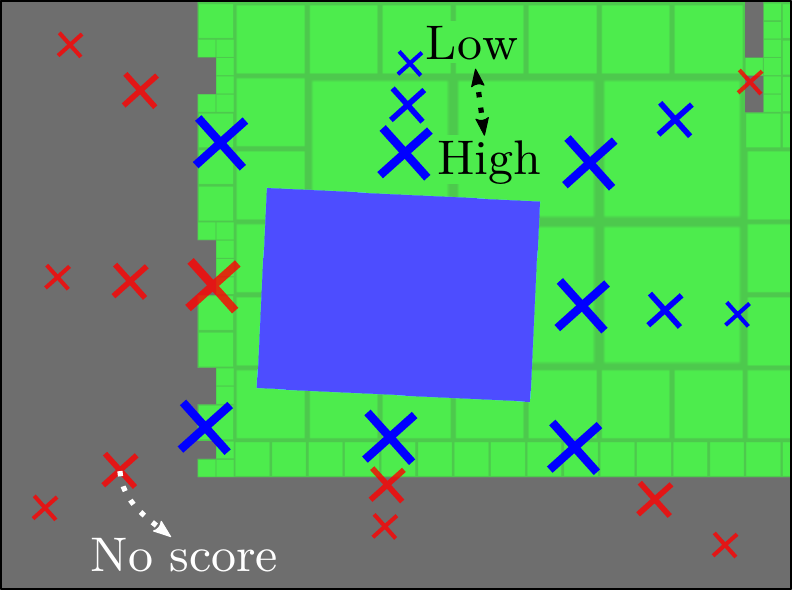} }
    \caption{These figures represent the main concept of our sampling-based footstep planner.
        (a) Our approach makes a footstep sample within the reachable region randomly and connects a valid sample to a support footstep.
        (b) The sampling process constructs a tree structure consisting of multiple footstep paths. This example shows two different paths colored by blue and red.
        (c) Our method computes the safety score using the multiple test points of a footstep. 
        The blue and red X marks represent the points that receives some scores and no score, 
       respectively.
        Different sizes of each mark indicate various weight values of the safety scores.
    }
    \label{fig:footstep_planning}
\end{figure*}
\section{Footstep Planning}

\newcommand{\YS}[1] {\textcolor{magenta}{YS: #1}}
\newcommand{\SB}[1] {\textcolor{blue}{SB: #1}}
\newcommand{\Skip}[1]{}





\begin{algorithm}[t]
	\caption{{\sc FOOTSTEP PLANNING} }
	\begin{algorithmic}[1]
		\renewcommand{\algorithmicrequire}{\textbf{Input:}}
		\renewcommand{\algorithmicensure}{\textbf{Output:}}
		\REQUIRE A steppable geometry $G$, a start footstep $q_{init}$
		\ENSURE A footstep path
		\STATE $T \gets \{q_{init}\}$
		\WHILE{termination condition is not satisfied}
			\STATE $q_{sup} \gets RandomSupportFootstep(T)$
			\STATE $q_{rand} \gets RandomFootstep(q_{sup})$
			\IF{$ValidityTest(q_{rand}, G)$}
				\STATE Connect $q_{sup}$ with $q_{rand}$
				\STATE Insert $q_{rand}$ to $T$
			\ENDIF
		\ENDWHILE
		\STATE $P \gets FootstepPathCandidates(T)$
		\RETURN $BestFootstepPath(P, G)$
	\end{algorithmic}
	\label{algo:footstep_planning}
\end{algorithm}

Utilizing the continuously updated terrain map, we generate feasible footsteps while preserving a real-time performance for re-planning. 
The vision sensor has a limited field of view, which causes a partial geometry of the environment.
Therefore, in the on-the-fly locomotion system, we adopt a local approach to plan a footstep path given the partially observed steppable objects.

In this work, a steppable region represents an area where the next footstep of a support footstep is feasible geometrically and kinematically as well. As shown in Fig.~\ref{fig:footstep_planning}-(a), we define the steppable region as the intersection of two parts: 1) geometric representations of steppable objects and 2) reachable region where the robot can take one step from the support footstep kinematically.
Our framework obtains the geometry of steppable objects from the vision system as a grid-based volumetric representation (Fig.~\ref{fig:mapping}). 
In the case of the reachable region, we use a kinematic condition to define the region of a support footstep throughout various experiments.
However, computing the explicit representation of the steppable region, intersection of steppable objects and the reachable region becomes an overhead in a real-time system.

We, therefore, exploit a sampling approach instead of finding the steppable region explicitly (Alg.~\ref{algo:footstep_planning}).
The proposed footstep planner samples a next footstep within the reachable region of the support footstep.
This sampling step makes a new, kinematically feasible footstep.
Our method then checks the geometric validity of the sampled footstep as shown in Fig.~\ref{fig:footstep_planning}-(a).
The volumetric representation of the footstep should be within the geometry of the steppable environment.
As a result of the test, a valid sample indicates that the footstep is feasible geometrically and kinematically.
Our approach connects the valid footstep to its support footstep but discards an invalid sample.
As shown in Fig.~\ref{fig:footstep_planning}-(b), repeating this random sampling process generates multiple footstep paths in a tree structure. 
Each path of the footstep tree consists of the sequential footsteps linked from a root to a leaf.
The proposed planner chooses one footstep path among the multiple candidates in consideration of an uncertainty issue.
The terrain map can have a discretization error of the grid-based representation, or the robot may not take the next footstep toward the planned pose exactly.
We design a safety score to deal with the problem, representing how a footstep is far from the terrain edges.
For example, two footstep paths shown in Fig.~\ref{fig:footstep_planning}-(b) have the different levels of safety scores.
The blue footstep path has a higher safety score than the red path because the footsteps of the blue path are located more tightly within the inner parts of the steppable geometry.

Specifically, our method selects the footstep path having the best safety score:
\begin{equation}
p^{*}=\argmax_{p \in P} \, f_{safety}(p),
\end{equation}
where $P=\{p_{1}, p_{2}, \dots, p_{N}\}$ indicates a set of the $N$ footstep paths that our planner generates.
We design the function $f_{safety}(p)$ that computes a safety score of the footstep path $p$ consisting of the $M$ footsteps, $p=\{q_{1}, q_{2}, \dots, q_{M}\}$:
\begin{equation}
f_{safety}(p)=\sum_{m=1}^{M} \sum_{k=1}^{K} w_{k}^{m} \, \mathbb{I}(\mathbf{x}_{k}^{m} \, \text{is onto} \, G),
\label{eq:safety_score_function}
\end{equation}
where $\mathbf{x}_{k}^{m}$ represents $k$-th test point of $m$-th footstep associated with the footstep path $p$.
Our method computes the safety of a footstep using its surrounding test points as shown in (Fig.~\ref{fig:footstep_planning}-(c)),
in order to efficiently check whether the footstep is located nearby the edges of steppable objects.
If a test point $\mathbf{x}_{k}^{m}$ is onto the steppable objects $G$, our planner makes a score as a weight $w_{k}^{m}$ depending on the distance between the center of footstep and the test point; e.g., we can use the radial basis function (RBF) kernel.
Otherwise, we give no score value at the test point.
Note that the indicator function $\mathbb{I}(\cdot)$ in Eq.~\ref{eq:safety_score_function} returns $1$ if the input state is true and $0$ if not. 
Based on the scoring policy, our method selects the final footstep path with the highest safety score, consisting of the geometry-aware safe footsteps.

\Skip{
Specifically, we check whether each surrounded test point of a footstep is within the steppable objects like Fig.~\ref{fig:footstep_planning}-(c).
If so, our planner computes different safety values depending on the distance between the footstep and the test point. 
On the other hand, we give no score value to the test point.
Using the scoring strategy, we can design the safety score of a footstep path 
as the summation of the values over all the test points associated with the footsteps.
Based on the scoring policy, our method selects the final footstep path with the highest safety score, consisting of the geometry-aware safe footsteps.
}

\section{Walking Pattern-Generator}

Given the collision-free and kinematically admissible geometric path, we generate CoM and foot trajectories over time that maintains stability of the compliant LIPM. The overall walking-pattern generator used for Gazelle platform is composed of CoM \& footstep trajectory generation and posture-stabilization-controller. Walking patterns are generated based on the robot model, and the controllers are constructed based on sensor feedback. This section will briefly introduce the walking pattern generation method and the required stabilization controller.

\subsection{CoM \& Footstep Trajectory}

The pattern-generator is extended upon Preview Control method~\cite{preview_control} in order to create robot's CoM trajectories from determined footsteps from our planner.  
For the preview control to be stable, the controller requires information of the next two footsteps in advance. Therefore, for dynamic walking, the planner is bounded to generate a minimum footstep array length of two.

To accommodate on-the-fly replanning of the landing footstep, we utilize an efficient vector-type struct called Step Data Buffer (SDB) to store current and future footstep positions and timing. The SDB can be dynamically modified any time during the gait cycle by the planner or balancing controller, even in middle of the swing-foot phase.This is enabled through our specific implementation of a short-cycle preview controller~\cite{fast_preview} to stably re-create CoM trajectories at control interval of 500Hz.

\begin{figure}[t]
    \centering
    \includegraphics[scale=0.45]{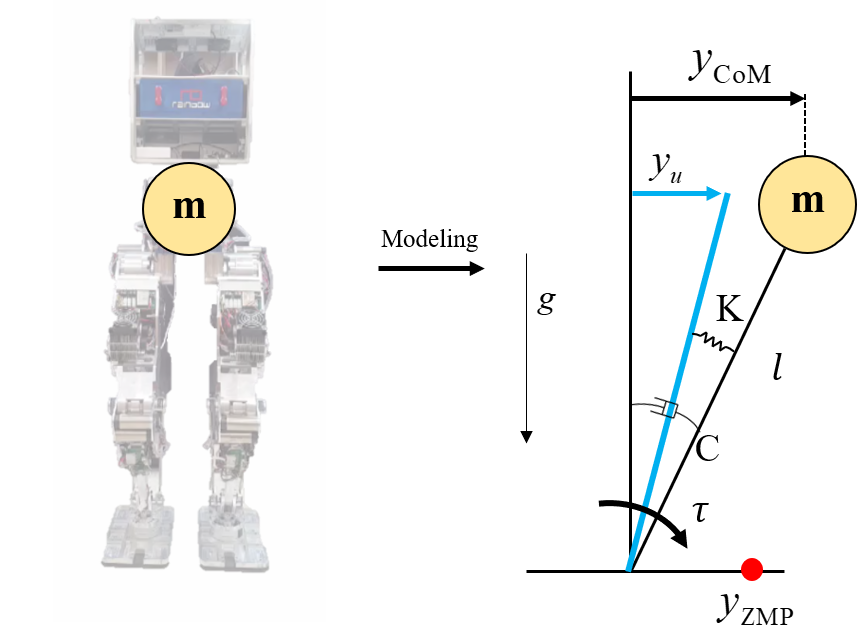}
    \caption{Compliant-LIPM modeling of real robot. In the real situation robot does not behave like ridged body due to joint, structure stiffness and compliance between the ground and the foot.}
    \label{fig:Compliant-LIPM}
\end{figure}


\begin{figure}[t]
    \centering
    \includegraphics[scale=0.365]{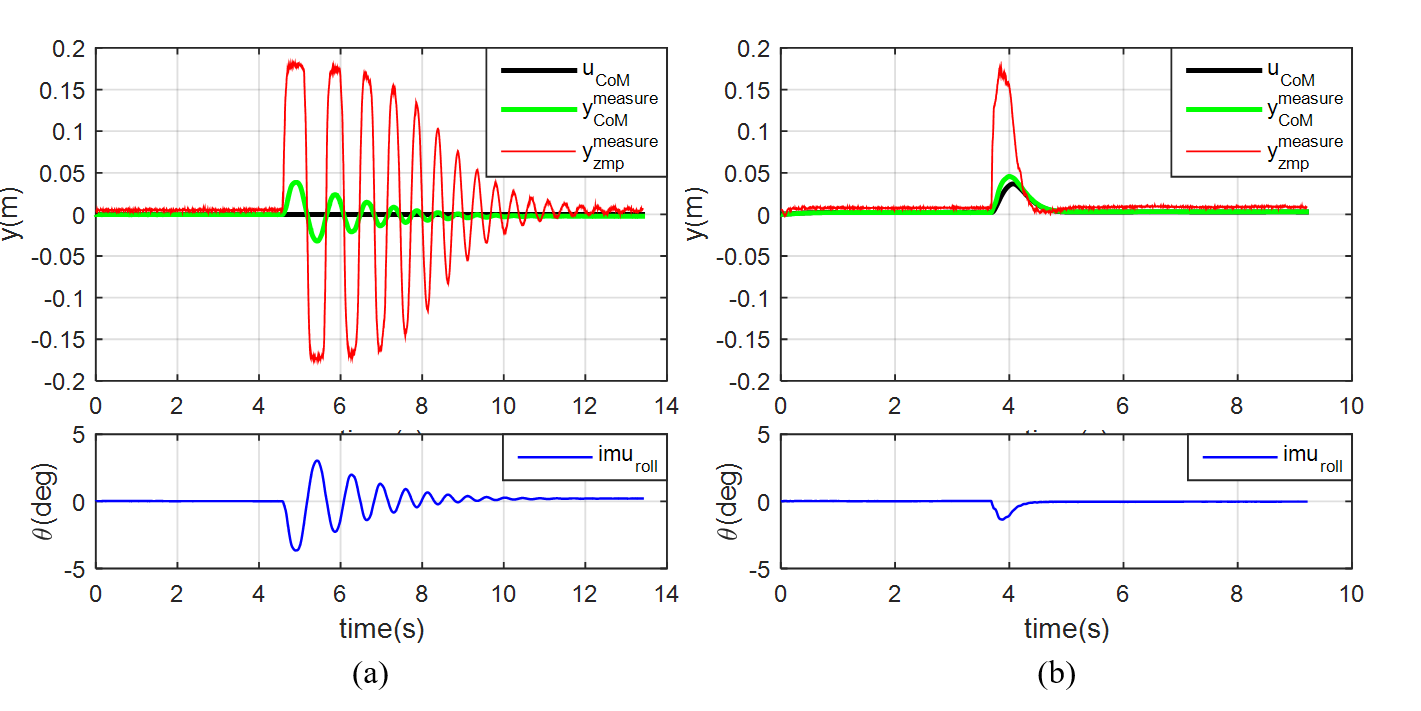}
    \caption{CoM damping control performance tested through robot's response to being pushed in the lateral direction. Graphs show the sensor values and the CoM position of the robot. (a) Without control, the body vibrates upon push because of high stiffness. (b) The damping control raises system damping and becomes robust to impact. }
    \label{fig:Damping_control_result}
\end{figure}

\subsection{Posture-Stabilization Controller}
Main purpose of posture stabilization control is to achieve stable walk according to the footstep information in SDB. Posture stabilization is implemented using contact force control and vibration damping control~\cite{gazelle_design} summarized below.

\subsubsection{Vibration Damping Control}
When a sudden external impact is applied to a standing robot, the robot will oscillate with a damped sinusoidal form because of the stiffness of each leg joint, including the stiffness of the structure, and the rubber pad attached under the feet. Such oscillation can negatively affect perception accuracy of visual sensors due to shaking motions and blurred images.
By walking, however, the robot receives such external impact continuously due to contact force between its landing foot and hard ground. Although the vibrations from these impact will damp out after a certain period of time, the damping is not enough to stabilize the robot during the continuous walking. Therefore, it requires to increase the damping and decrease the stiffness of the system. We used simple LIPM model for the walking pattern generation; here compliant-LIPM model is be applied for damping controller (Fig.~\ref{fig:Compliant-LIPM}).

From the state space form modeling of Compliant-LIPM that has input of $y_{u}$ and output of $y_{ZMP}$, the full state feedback controller can be constructed. The full state feedback is designed to specify new stiffness and damping of the system, and estimated states are used for feedback. Fig.~\ref{fig:Damping_control_result} presents the result of the vibration damping controller. Detailed description about modeling and state estimation are available in \cite{gazelle_design}.

\subsubsection{Contact Force Control}
Foot contact force control is necessary for the robot to walk on uneven ground or in the case of perception error of the ground height. Such instance results in unexpected, early landing or late landing of the stepping foot, 
causing the robot to fall without proper stabilization. To solve this problem, the reference force and torque of the foot should be obtained from the reference trajectory of the robot and the current state. In~\cite{hyobin_tro}, we proposed a method for generating the reference torque of a foot using capture point feedback. The error between the reference capture point and the measured capture point led us to obtain the desired ZMP (cZMP). This cZMP was used to generate the ankle torque reference.

Foot force control was performed in the vertical direction using leg length control. The cZMP from the capture point feedback was used to generate the reference vertical force of both feet. According to relative position of cZMP about foot position, the weight to be distributed to both feet is determined. This was then transformed to the desired length of both legs. Detailed description of ankle torque and force control are in \cite{gazelle_controller}.



\section{EXPERIMENTAL RESULTS}

\begin{figure}[t]
    \centering
    \includegraphics[scale=0.3]{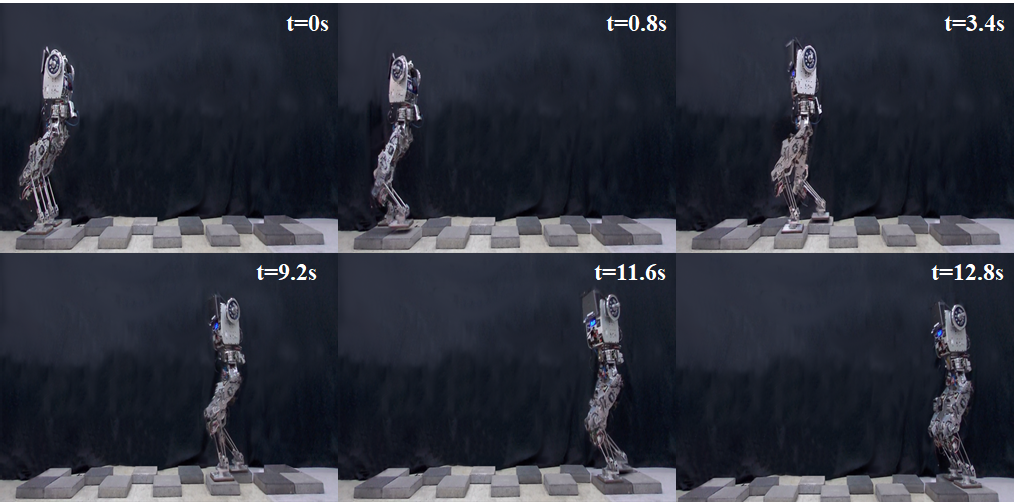}
    \caption{Terrain mapping, footstep planning, and trajectory generation pipeline integrated to walk continuously over uneven terrain at high speeds of 0.3 m/s.}
    \label{fig:SS_scenario_time}
\end{figure}

\begin{figure}[t]
    \centering
    \includegraphics[scale=0.6]{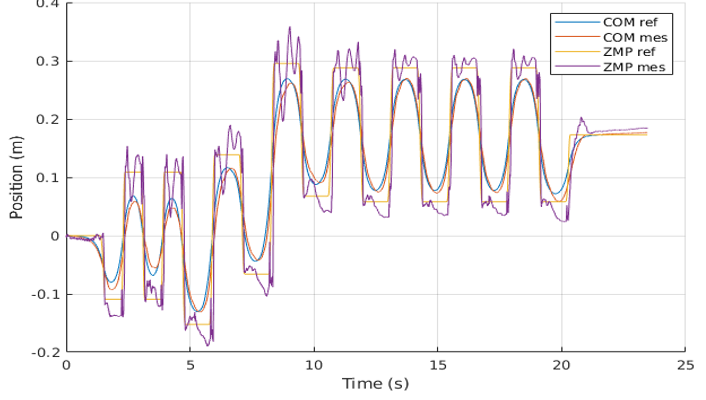}
    \caption{Measured ZMP and CoM tracks the desired trajectory created by the walking pattern generator resulting in stable balancing of the robot}
    \label{fig:SS_scenario_plot}
\end{figure}
We conduct various experiments to verify the performance of our streamlined perception-control pipeline. We first discuss experiments of the robot traversing scattered stepping stones for three meters without stopping as our validity test of the proposed system. We then present results of dynamic locomotion of the robot updating swing-leg trajectories on-the-fly as the terrain changes, and how the controller responds to stabilize the body. We lastly conduct experiments on a narrow path  the accuracy of the local mapping and footstep planner. 

As we focused on perception and footstep generation for dynamic locomotion, we did not implement a navigation strategy towards a specific goal, but rather embedded a forward-driven-bias into our planner to continuously generating steps until the robot reaches the end of the course.
The computation and performance, running on mini PC (6th gen i7 at 2.6GHz on 4 cores), of our proposed system can be summarized below. Note that these modules ran asynchronously on separate threads, and in the slowest-case scenario where each module is fed serially, the total pipeline computation is approximately 0.12 sec. Time for each module are as below:   

\begin{itemize}
    \item Depth Image acquisition: 33 msec (30 FPS, Wide FOV)
    \item Planar region segmentation and mapping: 67 msec (15Hz, 1cm resolution grid)
    \item Footstep planning: 5 msec (max 4 footsteps)
    \item Communication: 10 msec (ROS and PODO between 2 PCs) 
\end{itemize}

\subsection{System Performance Test}
The first experiment was traversing across three meters of scattered stepping stones displaced up to 0.3m apart for each footstep as seen in Fig.~\ref{fig:SS_scenario_time}. Using the proposed strategy, the robot successfully traversed across at a high speed of 0.3 m/s with 100\% success rate (tested 20 times). This default test scene was simplest compared to following experiments because the scene environment was static and thus did not need on-the-fly replanning of foot trajectories. In addition, the footstep positions generated from the planner were often located at the center of the stepping stone terrain, showing successful selection of path candidate with highest safety score as implemented. With planner that prioritized footstep positions at center of the terrain, traversing across stepping stones were more tolerant to perception error. For example, given the width of the stone and robot foot, the foot width covered only 50\% of the stone while the foot width covered 85\% of the narrow path below.    


\subsection{Dynamic Walking on Changing Terrain}
\begin{figure}[t]
    \centering
    \includegraphics[scale=0.35]{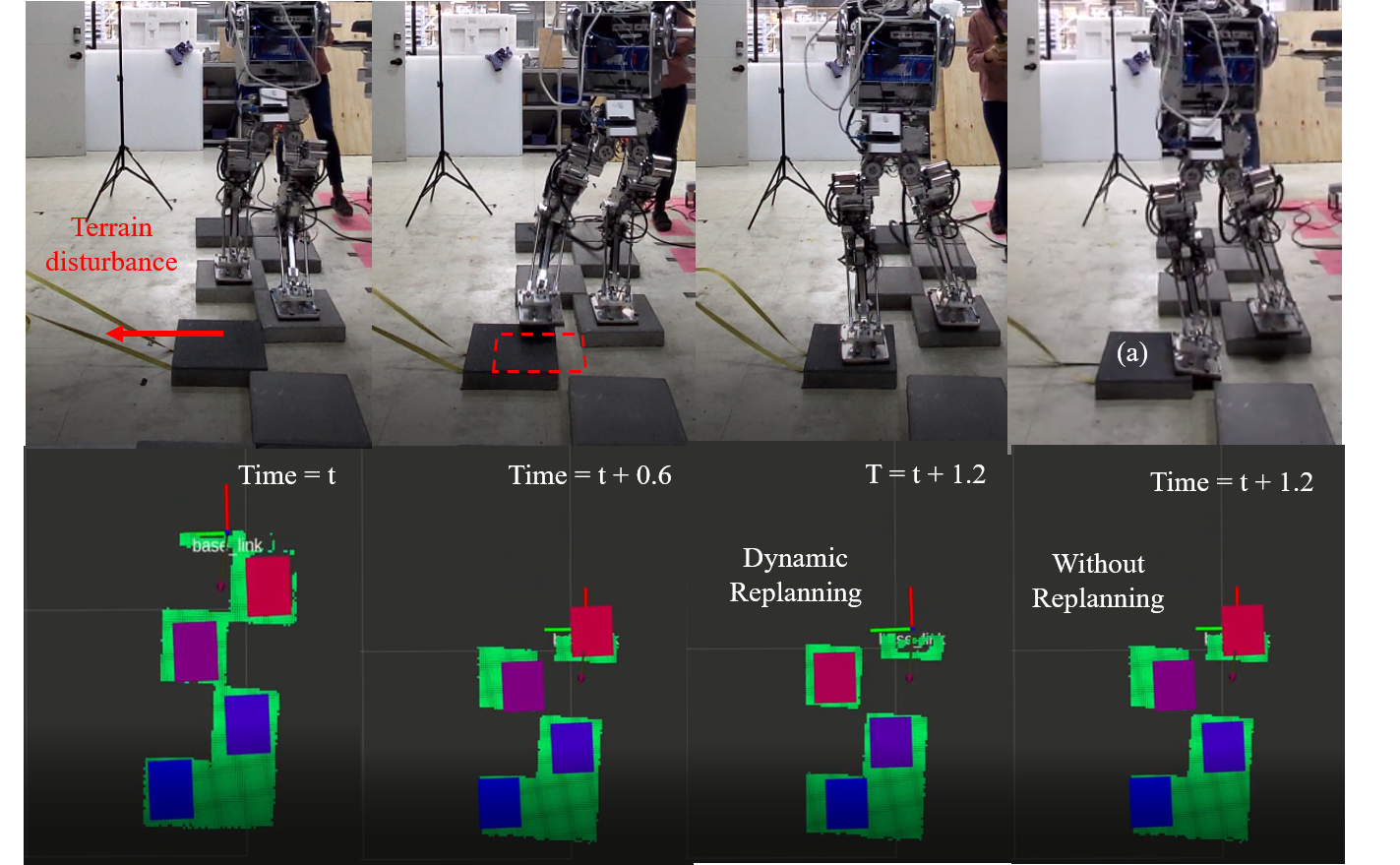}
    \caption{(a) Terrain disturbance is introduced during the robot's swing leg phase. Dynamic replanning 
   enables updated feasible footsteps on-the-fly. (b) Without replanning, the robot is susceptible to terrain disturbances and steps over the edge.}
    \label{fig:SS_dynamic_pic}
\end{figure}

\begin{figure}[t]
    \centering
    \includegraphics[scale=0.5]{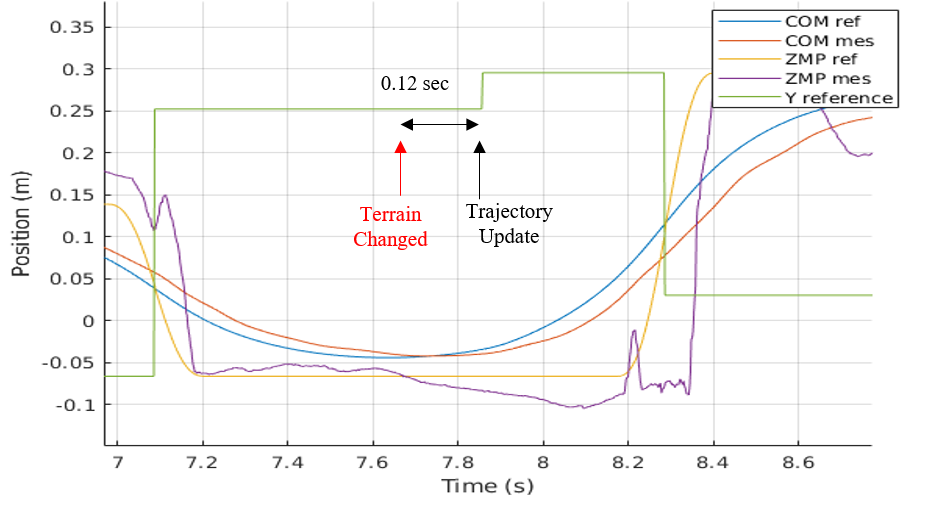}
    \caption{Trajectory of the swing-foot updated on-the-fly within 0.12 seconds after terrain change is introduced. ZMP and CoM trajectory is dynamically modified, yet is continuous and stable.}
    \label{fig:SS_dynamic_plot}
\end{figure}

As an extension to the previous test, the dynamic locomotion capability was tested by introducing a terrain disturbance as the robot is walking. During the robot's swing phase, the position of the stone where the immediate landing foot would have landed was changed approximately 8cm by pulling on a rope. As seen in Fig.~\ref{fig:SS_dynamic_pic} (a), during the short time-frame of the robot's swing phase, the pipeline was able to detect change in terrain, and updated its swing-leg trajectory under 0.12 second to readjust its landing footstep. Without such dynamic replanning capability, as shown in Fig.~\ref{fig:SS_dynamic_pic} (b), the robot would be susceptible to sudden terrain changes because it would not be able to update its landing footstep position on-the-fly, causing the robot to fall over. The upper constraint on how far the robot could readjust its swing-leg trajectory on-the-fly was experimentally determined to be approximately 0.5m euclidean distance from its previous footstep. For test values exceeding that bound, the robot would fall over in mid foot-swing in which the measured ZMP could not closely track the newly generated reframe ZMP. For sudden terrain change below 0.1m, the streamlined perception could generate newly feasible trajectory within 0.12sec, where the new CoM and ZMP reference would smoothly update with previous trajectory as shown in Fig.~\ref{fig:SS_dynamic_plot}.
As such, we set upper bound on how much the robot could adjust its landing foot position in mid-swing without falling over.    

\begin{figure}[h!]
    \centering
    \includegraphics[scale=0.3]{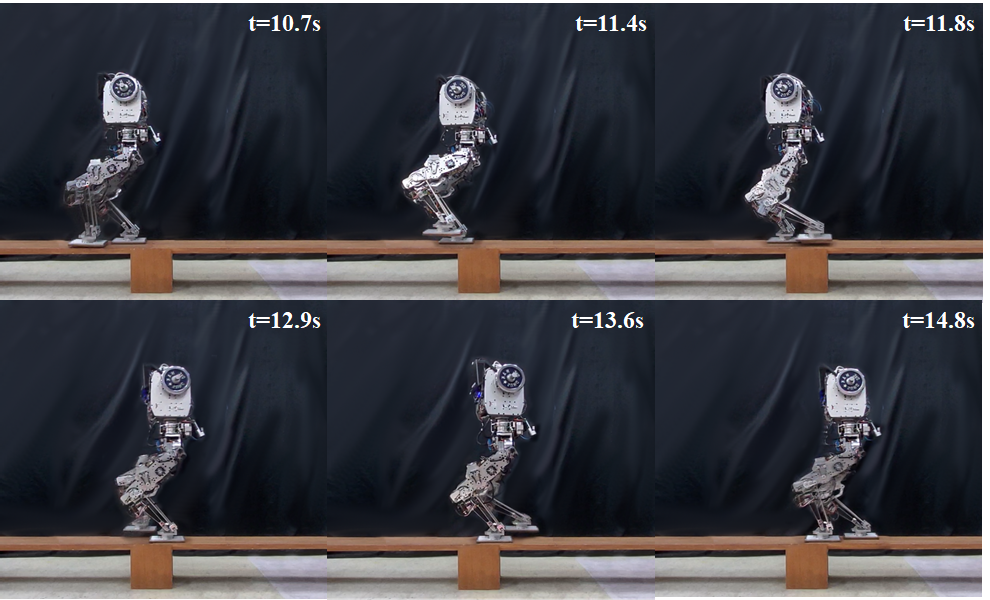}
    \caption{Walking continuously over narrow beam path with perception accuracy of 0.02m}
    \label{fig:NP_accuracy}
\end{figure}

\subsection{Precision Walking on Narrow Path}
In our final experiment, 
we verify the precision of our perception as well as our controller. To do so,  we set up an environment consisting of two platforms spanned by a narrow path of length three meters and width 0.3m, which is approximately half the width of the robot Fig.~\ref{fig:NP_accuracy}. In contrast to previous stepping stone walking pattern, the narrow path prevents the robot from placing both feet side-by-side, and constrains the robot to place one foot in front of the other. The small subset of kinematically reachable area and geometrically steppable region requires significant precision of the pipeline in order to not fall over the edge. To do so, the map was densely generated by increasing the resolution of the map to 0.5cm while narrowing the region of interest in octomap in order to maintain mapping rate. The robot successfully traversed narrow path with 90\% accuracy (tested 20 times), the two times failing due experimental mishandling where the robot's tethered power-cord got stuck on the narrow path setup.




\section{CONCLUSION}
In this paper we present an efficient geometric footstep planner and the corresponding walking controller that enables dynamic humanoid locomotion over uneven terrain. As dynamic locomotion, we refer first
to the continuous walking motion without stopping, and second
to the on-the-fly replanning of the landing footstep position in
middle of the swing phase during the robot gait cycle. 
This is mainly achieved through the streamlined integration between
an efficient sampling-based planner and robust walking controller. The footstep planner is able to generate feasible footsteps
within 5 milliseconds, and the controller is able to generate a
new corresponding swing leg trajectory as well as the whole-body motion to dynamically balance the robot to the newly updated footsteps.  
The proposed perception-control pipeline is evaluated and demonstrated with real experiments using a full-scale humanoid to traverse at high speeds of 0.3 m/s across uneven terrains featured by static stepping stones, dynamically movable stepping stone, and narrow path. Our current perception pipeline handles only uneven terrain with flat surfaces. For future work, we aim to cover terrains with varying heights and surface normal. We believe the current framework will extend nicely without accruing significant computation time.  


\section{ACKNOWLEDGEMENT}
This work was supported by the Technology Innovation Program (or Industrial Strategic Technology Development Program (0070171, Development of core technology for advanced locomotion/manipulation based on high-speed/power robot platform and robot intelligence) funded By the Ministry of Trade, industry \& Energy(MI, Korea).

\addtolength{\textheight}{-12cm}   



\bibliographystyle{./IEEEtran} 
\bibliography{./egbib2}

\begin{thebibliography}{10}
\providecommand{\url}[1]{#1}
\csname url@rmstyle\endcsname
\providecommand{\newblock}{\relax}
\providecommand{\bibinfo}[2]{#2}
\providecommand\BIBentrySTDinterwordspacing{\spaceskip=0pt\relax}
\providecommand\BIBentryALTinterwordstretchfactor{4}
\providecommand\BIBentryALTinterwordspacing{\spaceskip=\fontdimen2\font plus
\BIBentryALTinterwordstretchfactor\fontdimen3\font minus
  \fontdimen4\font\relax}
\providecommand\BIBforeignlanguage[2]{{%
\expandafter\ifx\csname l@#1\endcsname\relax
\typeout{** WARNING: IEEEtran.bst: No hyphenation pattern has been}%
\typeout{** loaded for the language `#1'. Using the pattern for}%
\typeout{** the default language instead.}%
\else
\language=\csname l@#1\endcsname
\fi
#2}}

\bibitem{asimo_cmu}
J.~Chestnutt, M.~Lau, G.~Cheung, J.~Kuffner, J.~Hodgins, and T.~Kanade,
  ``Footstep planning for the honda asimo humanoid,'' in \emph{IEEE
  International Conference on Robotics and Automation}, 2005.

\bibitem{2d_edge_detect}
M.~Asatani, S.~Sugimoto, and M.~Okutomi, ``Real-time step edge estimation using
  stereo images for biped robot,'' in \emph{IEEE International Conference on
  Intelligent Robots and Systems}, 2011.

\bibitem{2d_geometric_plan}
P.~Karkowski and M.~Bennewitz, ``Real-time footstep planning using a geometric
  approach,'' in \emph{IEEE International Conference on Robotics and
  Automation}, 2016.

\bibitem{mit_planning}
M.~Fallon, P.~Marion, R.~Deits, T.~Whelan, M.~Antone, J.~McDonald, and
  R.~Tedrake, ``Continuous humanoid locomotion over uneven terrain using stereo
  fusion,'' in \emph{IEEE International Conference on Humanoid Robots}, 2015.

\bibitem{ihmc_planning}
R.~Griffin, G.~Wiedebach, S.~McCrory, S.~Bertrand, I.~Lee, and J.~Pratt,
  ``Footstep planning for autonomous walking over rough terrain,'' in
  \emph{IEEE International Conference on Humanoid Robots}, 2019.

\bibitem{inaba_vision_recovery}
Y.~Kojio, Y.~Omori, K.~Kojima, F.~Sugai, Y.~Kakiuchi, K.~Okada, and M.~Inaba,
  ``Footstep modification including step time and angular momentum under
  disturbances on sparse footholds,'' in \emph{IEEE Robotics and Automation
  Letters, VOL. 5, NO. 3}, July 2020.

\bibitem{gazelle_design}
H.~Jeong, K.~K. Lee, W.~Kim, I.~Lee, , and J.-H. Oh, ``Design and control of
  the rapid legged platform gazelle,'' in \emph{Mechatronics 66, 102319},
  January 2020.

\bibitem{hyobin_tro}
H.~Jeong, I.~Lee, J.~Oh, K.~K. Lee, and J.-H. Oh, ``A robust walking controller
  based on online optimization of ankle, hip, and stepping strategies,'' in
  \emph{IEEE TRANSACTIONS ON ROBOTICS, VOL. 35, NO. 6}, December 2019.

\bibitem{gazelle_vision}
H.~Jeong, J.-H. Kim, O.~Sim, and J.-H. Oh, ``Avoiding obstacles during push
  recovery using real-time vision feedback,'' in \emph{IEEE International
  Conference on Intelligent Robots and Systems}, 2019.

\bibitem{Mhubo}
M.~Lee, Y.~Heo, J.~Park, H.~Yang, P.~Benz, H.~Jang, H.~Park, I.~Kweon, and
  J.~Oh, ``Fast perception, planning, and execution for a robotic butler:
  Wheeled humanoid m-hubo,'' in \emph{IEEE/RSJ International Conference on
  Intelligent Robots and Systems}, in press 2019.

\bibitem{fankhauser}
P.~Fankhauser, M.~Bloesch, and M.~Hutter, ``Probabilistic terrain mapping for
  mobile robots with uncertain localization,'' \emph{IEEE Robotics and
  Automation Letters}, vol.~3, no.~4, pp. 3019--3026, 2018.

\bibitem{preview_control}
S.~Kajita, F.~Kanehiro, K.~Kaneko, K.~Fujiwara, K.~Harada, K.~Yokoi, and
  H.~Hirukawa, ``Biped walking pattern generation by using preview control of
  zero-moment point,'' in \emph{IEEE International Conference on Robotics and
  Automation}, 2003.

\bibitem{fast_preview}
K.~Nishiwaki and S.~Kagami, ``High frequency walking pattern generation based
  on preview control of zmp,'' in \emph{IEEE International Conference on
  Robotics and Automation}, 2006.

\bibitem{gazelle_controller}
H.~Jeong, O.~Sim, H.~Bae, K.~Lee, J.~Oh, and J.-H. Oh, ``Design and control of
  the rapid legged platform gazelle,'' in \emph{EEE/RSJ International
  Conference on Intelligent Robots and Systems (IROS)}, 2017.

\end{thebibliography}

\end{document}